\documentclass[10pt,twocolumn,letterpaper]{article}

\usepackage{wacv}
\usepackage{times}
\usepackage{epsfig}
\usepackage{graphicx}
\usepackage{amsmath}
\usepackage{amssymb}
\usepackage{algorithm}
\usepackage{algpseudocode}
\usepackage{pdfpages}

\usepackage[pagebackref=true,breaklinks=true,letterpaper=true,colorlinks,bookmarks=false]{hyperref}

\wacvfinalcopy 

\def\wacvPaperID{66} 

\ifwacvfinal\fi
\begin{document}

\title{Product recognition in store shelves as a sub-graph isomorphism problem}

\author{Alessio Tonioni, Luigi Di Stefano \\
University of Bologna,\\
Department of Computer Science and Engineering (DISI)\\
Viale del Risorgimento 2, Bologna\\
{\tt\small $\lbrace$alessio.tonioni,luigi.distefano$\rbrace$@unibo.it }
}

\maketitle
\ifwacvfinal\thispagestyle{empty}\fi

\begin{abstract}
The arrangement of products in store shelves is carefully planned to maximize sales and keep customers happy. However, verifying compliance of real shelves to the ideal layout is a costly task routinely performed by the store personnel. In this paper, we propose a computer vision pipeline to recognize products on shelves and verify compliance to the planned layout. We deploy local invariant features together with a novel formulation of the product recognition problem as a sub-graph isomorphism between the items appearing in the given image and the ideal layout. This allows for auto-localizing the given image within the aisle or store and improving recognition dramatically.
\end{abstract}

\section{Introduction}
\label{sec:introduction}

Management of a grocery store or supermarket is a challenging task entailing personnel busy in supervising  shelves and the whole sale point. Technology advances may be deployed to provide more reliable information in real time to the store manager, so to coordinate human resources  more effectively.  Examples of tasks where innovation can improve current best practices are \textit{shelves analysis} (e.g. verifying low in stock or misplaced items), \textit{security} (e.g. reporting suspicious behaviours) and \textit{customer analysis} (e.g. analysing shopping patterns to improve  customer experience). However, a promising technological solution can be deployed in real shops as long as it turns out viable from a cost perspective, modifies current practices moderately and does not affect customer experience adversely. Computer vision techniques may fulfil the above requirements due to potential reliance on cheap cameras either mounted non-invasively in the store or embedded within the hand-held computers routinely used by sales clerks. 

The problem addressed in this paper is \textbf{visual shelf monitoring} through computer vision techniques. The arrangement of products in supermarket shelves is planned very carefully in order to maximize sales and keep customers happy. Shelves void, low in stock or misplaced products render it difficult for the customer to buy what she/he needs, which, in turn, not only leads to unhappy shoppers but also to significant loss of sales; as pointed out in \cite{gruen2002retail}, 31\% of customers facing a void shelf purchase the item elsewhere and 11\% do not buy it at all. The planned layout of products within shelves is called \textit{planogram}: it specifies where each product should be placed within shelves and how many \textit{facings} it should cover, that is how many packages of the same product should be visible in the front row of the shelf. Keeping shelves full as well as compliant to the planogram is a fundamental task for all types of stores that could lead to 7.8\% sales increase and 8.1\% profit improvement in just two weeks \cite{shapiro2009planogram}. However, thus far, planogram compliance is pursued by having sales clerks visually inspecting aisles during the quieter hours of the day. 

Computer vision may help to automate, at least partially, this task. As vouched by recently published patents \cite{larsen2013automated}, \cite{opalach2012planogram} and journal articles \cite{marder2015using}, some major corporations are currently investigating on deployment of state of the art computer vision techniques to pursue planogram compliance, with smaller emerging companies (such as \textit{Planorama} , \textit{Vispera}, \textit{Simble Robotics}) \footnote{\url{http://www.planorama.com/}, \url{http://vispera.co/}, \url{http://www.simberobotics.com/}} advertising this type of service alike. From a scientific perspective, attaining planogram compliance by automated visual analysis represents a very challenging task due to the large number of object instances that should be identified and localized in each scene, the presence of many distractors, the small differences between different instances of products belonging to the same brand and the varying lighting conditions. Accordingly, to the best of our knowledge, established scientific approaches have not emerged yet while industrial solutions seem either at a prototype stage or in the very early part of their life cycle.

In this paper we propose a computer vision pipeline that, given the planogram and an image of the observed shelves, can correctly localize each product, check whether the real arrangement is compliant to the planned one and detect missing or misplaced items. Key to our approach is a novel formulation of the problem as a sub-graph isomorphism between the product detected in the given image and those that should ideally be found therein given the planogram.  Accordingly, our pipeline relies on a standard feature-based object recognition step, followed by the novel graph-based consistency check and a final localized image search to improve the overall product recognition rate.

\section{Related Work}
\label{sec:statoArte}


The problem of automatically recognizing grocery products from images may in principle be traced back to the more general and extensively investigated subject of visual object recognition. However, as pointed out by Merler et al. \cite{merler2007recognizing}, dealing with grocery products on shelves exhibits peculiarities that render the task particularly challenging. Indeed, as also exemplified in the leftmost column of \autoref{fig:pipeline}, one has to rely on a single or a few views either synthetic (graphic renderings of the package) or taken in ideal studio-like conditions in order to model each product instance which, then, must be sought within images acquired in real settings. The scarcity and diversity of model images make it awkward to deploy directly object recognition methods, such as deep convolutional neural networks, that demand a large corpus of labeled training examples representative of unseen data. As noticeable in \autoref{fig:pipeline}, verifying planogram compliance calls for detecting and localizing each individual product instance within a shelves image crowded with lots of objects, some remarkably similar one to another. Moreover, the scene usually include several distractor items, such as vividly colored banner ads designed to attract customer eyes, that may mislead computer vision algorithms. As the operational conditions should  be left as unconstrained as possible, recognition algorithms should withstand working images featuring dramatic changes in color, intensity and even resolution, due to varying lighting conditions as well as deployment of diverse acquisition devices. 
In their work, Merler et al. \cite{merler2007recognizing} discuss the above-mentioned issues, propose a public dataset and pursue product recognition to realize an assistive tool for visually impaired customers. They assume that no information concerning product layout may be deployed to ease detection. Given these settings, the performance of the proposed systems turned out quite unsatisfactory in terms of both precision and efficiency. 

Further research has then been undertaken to ameliorate the performance of automatic visual recognition of grocery products  \cite{winlock2010toward}, \cite{varol2015toward}, \cite{cotter2014hardware}. In particular, Cotter et al.  \cite{cotter2014hardware}  report significant performance improvements by leveraging on machine learning techniques, such as HMAX and ESVM, together with HOG-like features. Yet, their proposal requires many training images for each product, which is unlikely feasible in real settings, and deploys a large ensemble of example-specific detectors, which makes the pipeline rather slow at test time. Moreover, adding a new type of sought product is rather cumbersome as it involves training a specific detector for each exemplar image, thereby also further slowing down the whole system at test time. 
The approach proposed in  \cite{cotter2014hardware} was then extended in \cite{advani2015visual} through a contextual correlation graph between products. Such a structure can be queried at test time to predict the products more likely to be seen given the last $k$ detections, thereby reducing the number of ESVM computed at test time and speeding up the whole system.  
 
Another relevant work is due to George at al. \cite{george2014recognizing}.  First, to reduce the search space of the actual detection phase, they carry out an initial classification to infer the categories of observed items.  Then, following detection, they run an optimization step based on a genetic algorithm  to detect the most likely products from a series of proposals. Despite the quite complex pipeline, when relying on only one model image per product the overall precision of the system is  below 30\%. The paper proposes also a publicly available dataset, referred to as  \textit{Grocery Products}, comprising 8350 product images classified into 80 hierarchical categories together with 680 high resolution images of shelves. In this paper, we use part of this public dataset as the main test bench for our method. 

Marder et al. in \cite{marder2015using} addressed our exact same problem of checking planogram compliance through computer vision. Their approach relies on detecting and matching SURF features \cite{bay2006surf} followed by visual and logical disambiguation between similar products. The paper reports a good 87.4\% product recognition rate on a publicly unavailable dataset of cereal boxes and hair care products, though precision figures are not highlighted. Their dataset includes 240 images with 980 instances of 223 different products, that is, on average $\approx4$ instances per image. To improve product recognition the authors deploy information dealing with the known product arrangement through specific hand-crafted rules, such as e.g. `conditioners are placed on the right of shampoos`. Differently, we propose to deploy automatically these kind of constraints by modeling the problem as a sub-graph isomorphism between the items detected in the given image and the planogram. Unlike ours, their method mandates a-priori categorization of the sought products into subsets of visually similar items. 

Systems to tackle the planogram compliance problem are described also in \cite{frontoni2014information}, \cite{frontoni2015embedded} and \cite{mankodiya2012challenges}. These papers delineate solutions relying either on large sensor/camera networks or mobile robots monitoring shelves while patrolling aisles. In contrast, our proposal would require just an off-the-shelf device, such as a smartphone, tablet or hand-held computer. 

\section{Proposed Pipeline}
\label{sec:system}

\begin{figure*}[t]
\centering
\includegraphics[width=150mm]{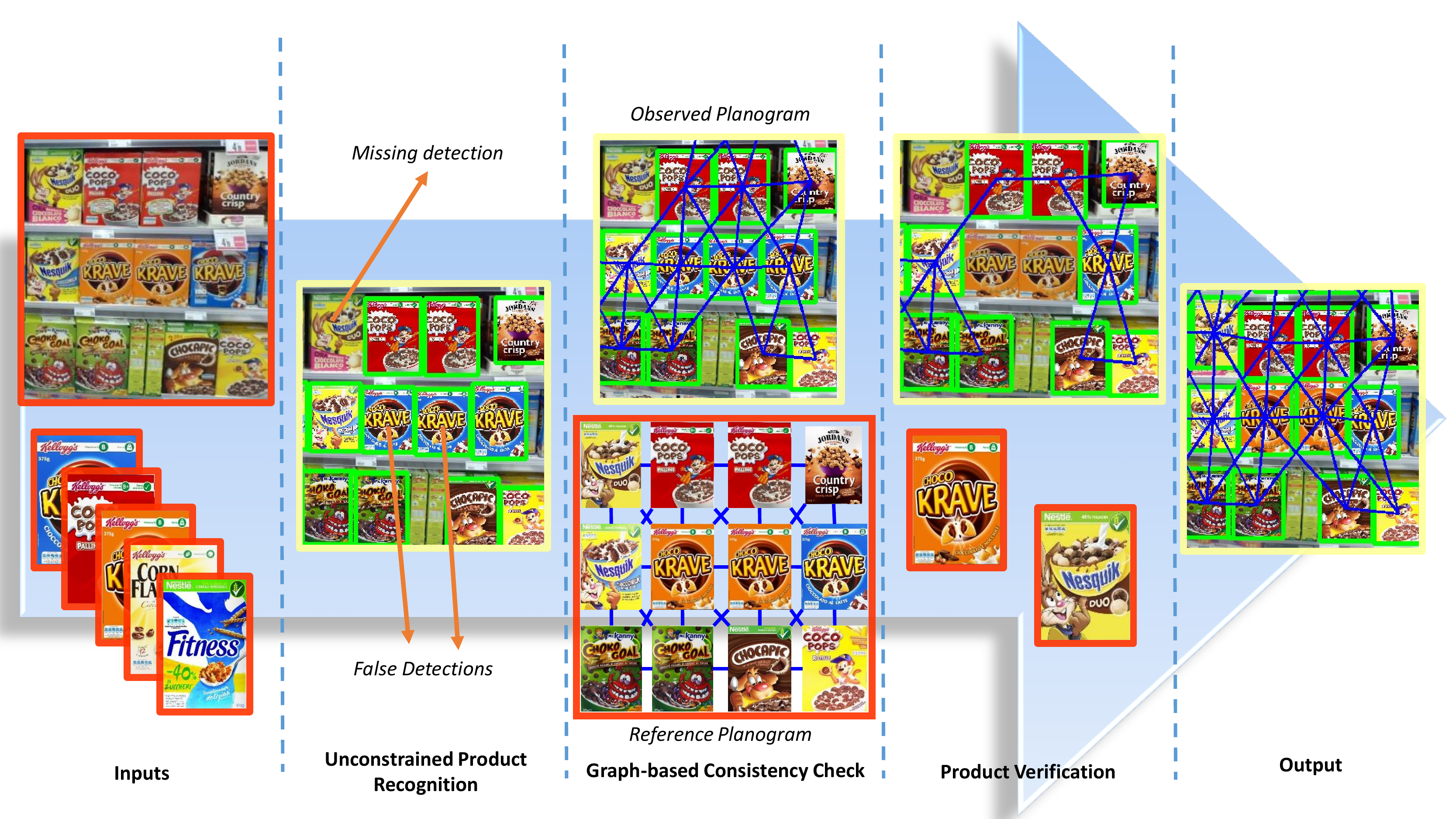} 
\caption{Overview of our pipeline. For each step we highlight the inputs and outputs through red and yellow boxes, respectively. Product detections throughout stages are highlighted by green boxes, while blue lines show the edges between nodes in both the \textit{Reference} and \textit{Observed} planograms.}
\label{fig:pipeline}
\end{figure*}

We address the typical industrial settings in which at least one model image per product together with a general schema of the correct disposition of items (the planogram) are available. At test time, given one image featuring products on shelves, the system would detect and localize each item and check if the observed product layout is compliant to the given planogram. As depicted in \autoref{fig:pipeline}, we propose to accomplish the above tasks by a visual analysis pipeline consisting of three steps. We provide here an overview of the functions performed by the three steps, which are described in more detail in the following Sub-sections.

The first step operates only on model images and the given shelves image. Indeed, to pursue seamless integration with existing procedures, we assume that the information concerning which portion of the aisle is observed is not available together with the input image. Accordingly, the first step cannot deploy any constraint dealing with the expected product disposition, and is thus  referred to as \textbf{Unconstrained Product Recognition}. 
As most product packages consist of richly textured piecewise planar surfaces, we obtained promising result through a standard object recognition pipeline based on local invariant features (as described, e.g., in \cite{lowe2004distinctive}). Yet, the previously highlighted nuisances cause both missing product items as well as false detections due to similar products. Nonetheless, the first step can gather enough correct detections to allow the successive steps to identify the observed portion of the aisle in order to deploy constraints on the expected product layout and improve product recognition dramatically. The output of the first step consists in a set of bounding boxes corresponding to detected product instances (see \autoref{fig:pipeline}).

From the second step, dubbed  \textbf{Graph-based Consistency Check},  we start leveraging on the information about products and their relative disposition contained in planograms. We choose to represent a planogram as a \textit{grid-like fully connected graph} where each node corresponds to a product facing and is linked to at most 8 neighbors at 1-edge distance, i.e. the closest facings along the cardinal directions. 
We rely on a graph instead of a rigid grid to allow for a more flexible representation; an edge between two nodes does not represent a perfect alignment between them but just proximity along that direction.

This abstract representation, referred to as \textit{Reference Planogram}, encodes information about the number of facings related to each product and the items placed close together in shelves.   
An example of \textit{Reference Planogram} is shown in \autoref{fig:pipeline}. 
The detections provided by the first step are used in the second to build automatically another \textit{grid-like graph} having the same structure as the \textit{Reference Planogram} and referred to as \textit{Observed Planogram}. 
Then, we find the \textit{sub-graph isomorphism} between the \textit{Observed} and  \textit{Reference} planograms, so as to identify local clusters of self-consistent detected products, e.g. sets of products placed in the same relative position in both the \textit{Observed} and \textit{Reference} planograms.
As a result, the second step ablates away inconsistent nodes from the \textit{Observed Planogram}, which typically correspond to false detections yielded by the first step. It is worth pointing out that, as the \textit{Observed Planogram} concerns the shelves seen in the current image while the \textit{Reference Planogram} models the whole aisle, matching the former into the latter implies localizing the observed scene within the aisle\footnote{More generally, matching the \textit{Observed} to a set of \textit{Reference} planograms does localize seamlessly the scene within a set of aisles or, even, the whole store.}. 

After the second step the \textit{Observed Planogram} should contain true detections only. Hence, those nodes that are missing compared to the \textit{Reference Planogram} highlight items that appear to be missing  wrt the planned product  layout.  The task of the third step, referred to as \textbf{Product Verification}, is to verify whether these product items are really missing in the scene or not. More precisely, we start considering the missing node showing the highest number of already assigned neighbors, for which we can most reliably determine a good approximation of the expected position in the image. Accordingly, a simpler computer vision problem than in the first step needs to be tackled, i.e. verifying whether or not a known object is present in a well defined ROI (Region of Interest) within the image.
Should the verification process highlight the presence of the product, the corresponding node would be added to the \textit{Observed Planogram}, so to provide new constraints between found items; otherwise, a planogram compliance issue related to the checked node is reported (i.e. missing/misplaced item). The process is iterated till all the facings in the observed shelves are either associated with detected instances or flagged as  compliance issues.

\subsection{Unconstrained Product Recognition}
\label{ss:ObjRecognition}

As already mentioned, we rely on the classical multi-object and multi-instance object recognition pipeline based on local invariant features presented in \cite{lowe2004distinctive}, which is effective with planar textured surfaces and scales well to database comprising several hundreds or a few thousands models, i.e. in the order of the number of different products typically sold in grocery stores and supermarkets. Accordingly, we proceed through feature detection, description and matching, then cast votes into a pose space by a Generalized Hough Transform 
that can handle multiple peaks associated with different instances of the same model 
in order to cluster correspondences and filter out outliers. In our settings, it turns out reasonable to assume the input image to represent an approximately frontal view of shelves, so that both in-plane and out-of-plane image rotations are small. Therefore, we estimate a 3 DOF pose (image translation and scale change). 

Since the introduction of SIFT \cite{lowe2004distinctive}, a plethora of other feature detectors and descriptors have been proposed in literature. Interestingly, the object recognition pipeline we used that is described in \cite{lowe2004distinctive} may be deployed seamlessly with most such newer proposals. Moreover, it turns out just as straightforward to rely on multiple types of features jointly to pursue higher sensitivity thanks to detection of diverse image structures.
Purposely, our implementation of the standard object recognition pipeline can run in parallel several detection/description/matching processes based on different features and have them eventually cast vote altogether within the same pose space. As reported in \autoref{sec:test}, we have carried out an extensive experimental investigation to establish which features would yield the best performance. 

\subsection{Graph-based Consistency Check}
\label{sec:graph_consistency}

To build the \textit{Observed Planogram} we instantiate a node for each item detected in the previous step and perform a loop over all detections to seek for bounding boxes around other detected items that are located close the current one. For each node, the search is performed along 8 cardinal directions (N, S, E, W, NW, NE, SW, SE) and, if another bounding box is found at a distance less than a dynamically determined threshold, an edge is created between the two nodes. In the given node the edge is labeled according to the search direction (e.g. N), oppositely in the found neighbor node (i.e. S). The graph is kept self-coherent, e.g. if node B is the \textit{North} node of A, then A must be the \textit{South} node of B. In case of ambiguity, e.g.  both  A and C found to be the \textit{South} node of B, we retain the edge between the two closest bounding boxes only. 

Once built, we compare the \textit{Observed} to the \textit{Reference Planogram} so to determine whether and how the two graphs overlap one to another. In theoretical computer science this problem is referred to as \textit{subgraph isomorphism} and  known to be NP-complete\cite{wegener2005complexity}. A general formulation may read as follows:  given two graphs $G$ an $H$, determine whether $G$ contains a subgraph for which does exist a bijection between the vertex sets of $G$ and $H$. However, given our strongly bounded graphs, we choose not to rely on one of the many general algorithms, like  e.g.  \cite{ullmann2010bit}, and, instead, devised an ad hoc heuristic algorithm that, casting ours as a constraint satisfaction problem, works fairly well in practice.

We formulate our problem as follows: given two graphs $I$ (\textit{Reference Planogram}) and $O$ (\textit{Observed Planogram}), find an isomorphism between a subset of nodes in $I$ and a subset of nodes in $O$ such that the former subset has the maximum feasible cardinality given product placements constraints. Each node in $I$ can be associated with a node in $O$ only if they both refer to the same product instance and exhibit coherent neighbors. In other words, we find the maximum set of nodes in graph $O$ that turn out self-consistent, i.e. their relative positions are the same as in the reference graph $I$.  

\begin{algorithm} 
\caption{Find \textit{sub-graph isomorphism} between $I$ and $O$}
\label{cod:algo}
\begin{algorithmic}
	\State $C_{max} \gets 0$
	\State $S_{best} \gets \emptyset$
	\State $\mathcal{H} \gets CreateHypotheses(\textit{I,O})$
	\While {$\mathcal{H} \neq \emptyset$} 
		\State $ C, \mathcal{S}, h_{0} \gets FindSolution(\mathcal{H}, C_{max}, \tau)$
		\If { $C > C_{max}$}
			\State $S_{best},C_{max} \gets S,C$
		\EndIf
		\State $\mathcal{H} \gets \mathcal{H} - h_0$
	\EndWhile
	\State $ \textbf{return } S_{best}, C_{max}$
\end{algorithmic}
\end{algorithm}

\begin{figure}[t]
\centering
\includegraphics[width=85mm]{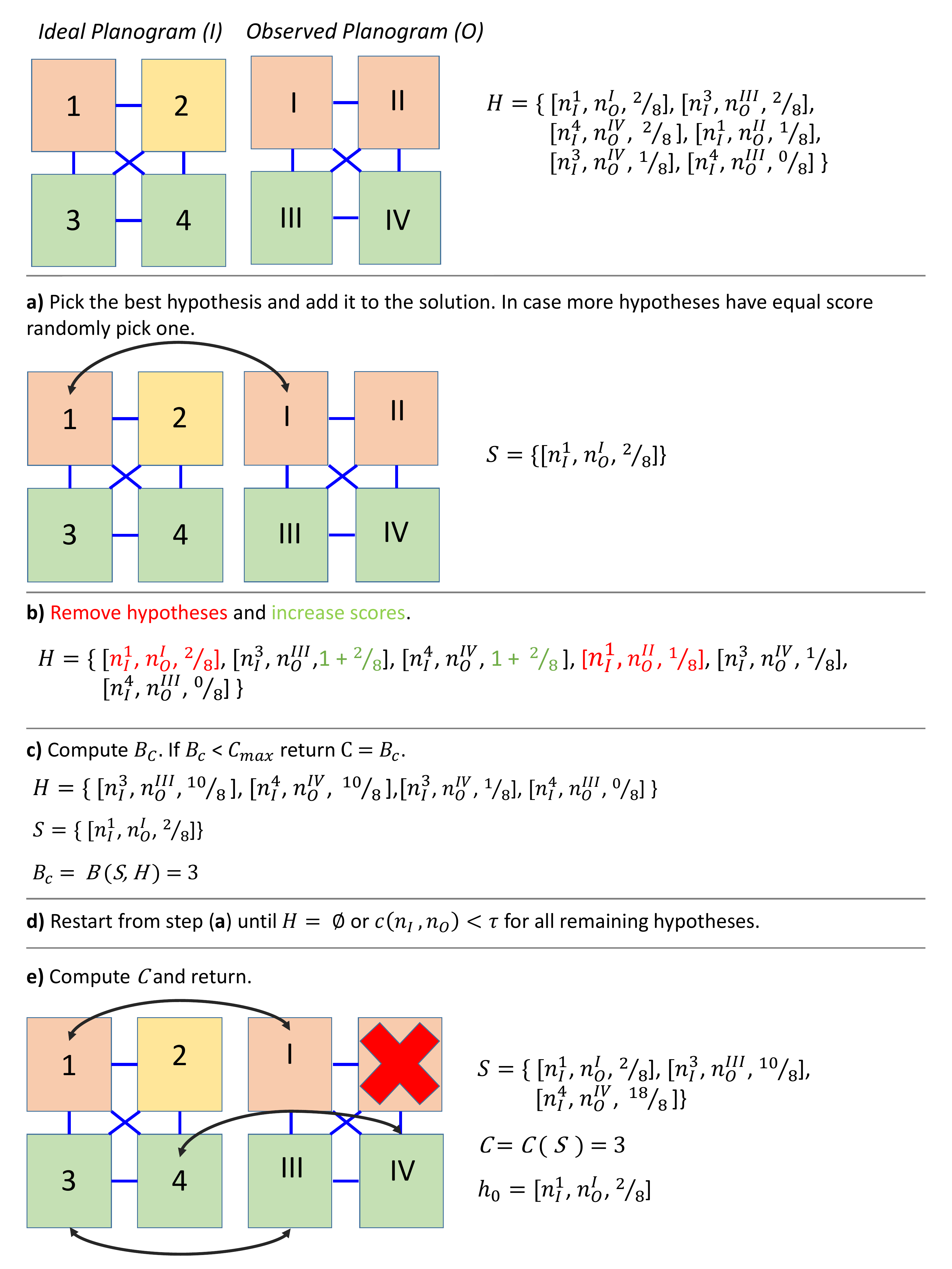} 
\caption{Toy example concerning two small graphs with 4 nodes used to describe procedure \textit{FindSolution}. The color of each node denotes the product the numbers within squares identify the different nodes in the text.}
\label{fig:find_solution}
\end{figure}

As illustrated in Algorithm \autoref{cod:algo}, the process starts with procedure \textit{CreateHypotheses}, which establishes  the initial set of hypotheses, $\mathcal{H}=\{\ldots h_{i} \ldots \}$,  $h_{i}=\{n_I,n_O,c(n_I,n_O)\}$, with $n_I$ and $n_O$ denoting, respectively, a node in $I$ and  $O$ related to the same product and
$c(n_I,n_O)=\frac{nn_c}{nn_t}$ with $nn_c$ number of coherent neighbors (e.g. refering to the same product both in $O$ and $I$) and $nn_t$ number of neighbors for that node in $I$. 
\textit{CreateHypotheses} iterates over all $n_I\in I$ so to instantiate all possible hypotheses. An example of the hypotheses set determined by \textit{CreateHypotheses} given $I$ and $O$ is shown in the first row of \autoref{fig:find_solution}. Then, procedure \textit{FindSolution} finds a solution, $\mathcal{S}$, by iteratively picking the hypothesis featuring the highest score. The first hypothesis picked in the considered example is shown in \autoref{fig:find_solution}-a). Successively,  $\mathcal{H}$ is  updated by removing the hypotheses containing either of the two nodes in the best hypothesis and increasing the scores of hypotheses associated with coherent neighbors (\autoref{fig:find_solution}-b)). Procedure \textit{FindSolution} returns also a confidence score for the current solution, $C$, which takes into account the cardinality of $\mathcal{S}$, together with a factor which penalizes the presence in  $O$ of disconnected sub-graphs that exhibit relative distances different than those expected given the structure of $I$\footnote{In the toy example in \autoref{fig:find_solution}, $O$ does not contain disconnected sub-graphs.} which instead is always fully connected. \textit{FindSolution} takes as input the score of the current best solution, $C_{max}$, and relies on a branch-and-bound scheme to accelerate the computation. In particular, as illustrated in \autoref{fig:find_solution}-c), after updating $\mathcal{H}$ (\autoref{fig:find_solution}-b)), \textit{FindSolution}  calculates an upper-bound for the score, $B_C$, by adding to the cardinality of $\mathcal{S}$ the number of hypotheses in $\mathcal{H}$ that are not mutually exclusive, so as to early terminate the computation when the current solution can not improve $C_{max}$.  
The iterative process continues with picking the new best hypothesis until $\mathcal{H}$ is found empty or containing only hypotheses with confidence lower then a certain threshold $\tau$ (\autoref{fig:find_solution}-d). The found solution, $\mathcal{S}$, contains all the hypotheses that are self-consistent and such that each node $n_I$ is either associated with a node $n_O$ or to none, as shown in the last row of \autoref{fig:find_solution}.
In the last step ((\autoref{fig:find_solution}-e)), the procedure computes $C$ and returns also the first hypothesis, $h_0$, that was added into $\mathcal{S}$, i.e. that with the highest score $c(n_I,n_O)$ (\autoref{fig:find_solution}.-a)). Upon returning from \textit{FindSolution}, the algorithm checks whether or not the new solution  $\mathcal{S}$ improves the best one found so far and removes $h_{0}$ from $\mathcal{H}$ (see Algorithm \autoref{cod:algo}) to allow evaluation of another solution based on a different initial hypotheses.

As a result, Algorithm \autoref{cod:algo} finds self-consistent nodes in $O$ given $I$, thereby removing inconsistent (i.e. likely false) detections and localizing the observed image wrt to the planogram. Accordingly, the output of the second steps contains information about which items appear to be missing given the planned product layout and where they ought to be located within the image. 

\subsection{Product Verification}
\label{sec:localSearch}

\begin{figure}[t]
\centering
\includegraphics[width=80mm]{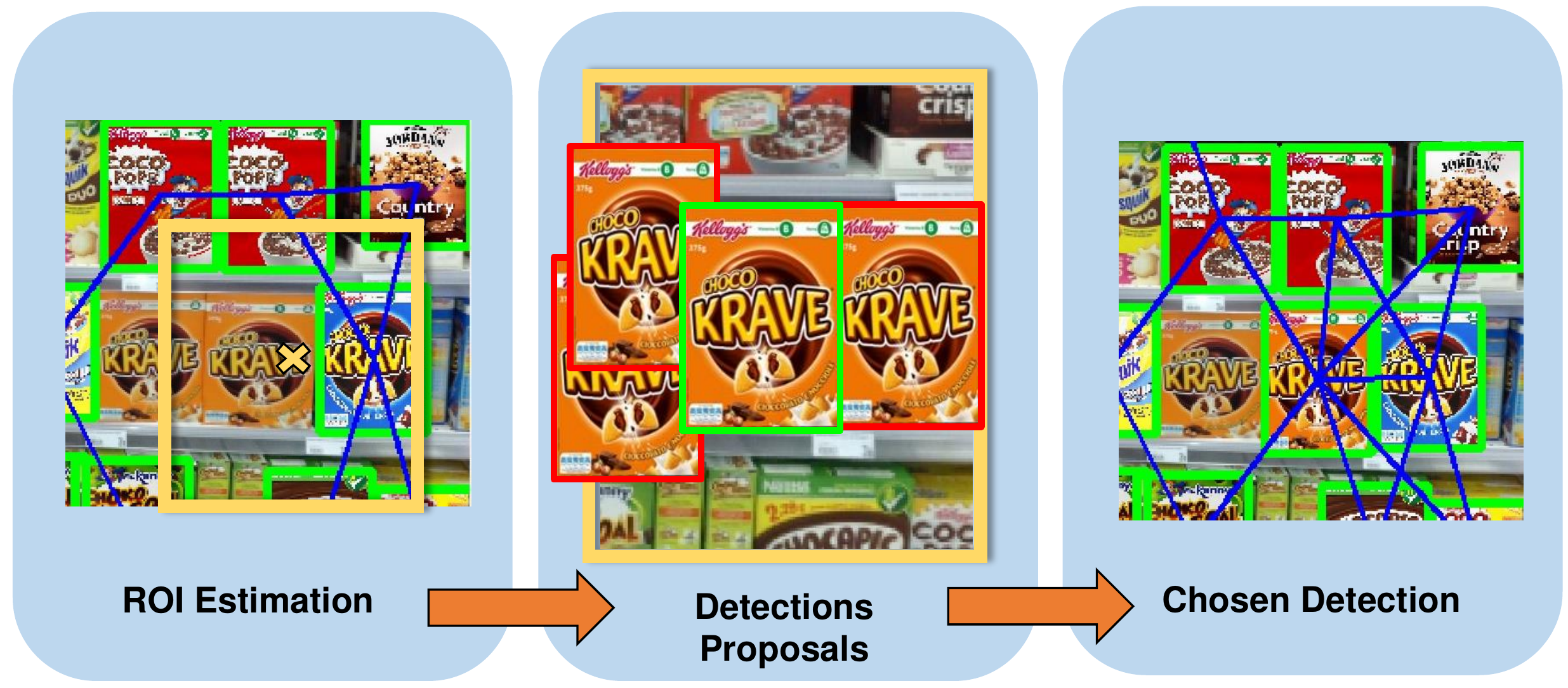} 
\caption{One iteration of the Product Verification. The estimated ROI is drawn in yellow. The correct proposal is highlighted in green while others are in red.}
\label{fig:step3}
\end{figure}

We use an iterative procedure whereby each iteration tries to fill the observed planogram with one seemingly missing object. As illustrated in \autoref{fig:step3}, each iteration proceeds through three stages. We start with the missing element featuring the highest number of already detected neighbors. The positions of these neighbors provide clues on where the missing product should appear in the image. In particular, the position and size of each neighbor, together with the average edge length in the \textit{Observed Planogram}, provide an estimation of the center of the missing element: averaging estimations across the neighbors yields a good approximate position.
Then, we define a coarse image ROI centered at this position by estimating the size of the missing element\footnote{Store databases contain product sizes: the image size of a missing product can be estimated from those of the detected neighbors and the known metric sizes.} and allowing for some margin on account of possible localization inaccuracies.  

Given the estimated ROI, the second stage attempts to find and localize the missing product therein. As already pointed out, unlike the initial step of our pipeline, here we now know exactly which product is sought as well as its approximate location in the image. To look for the sought product within the ROI, we have experimented with template matching techniques as well as with a similar pipeline based on local features as deployed for Unconstrained Product Recognition (\autoref{ss:ObjRecognition}). The latter, in turn, would favorably reuse the image features already computed within the ROI in the first step of our pipeline, so as to pursue matching versus the features associated with the model image of the sought product only and, accordingly, cast votes in the pose space. Both approaches would provide a series of Detection Proposals (see \autoref{fig:step3}).

Detection proposals are analyzed in the last stage of an iteration by first discarding those featuring bounding boxes that overlap with already detected items and then scoring the remaining ones according to the coherence of the position within the (\textit{Observed Planogram}) and the detection confidence. As for the first contribution to the score,  we take into account the error between the center of the proposal and that of the ROI estimated in the first stage (so to favor proposals closer to the approximated position inferred from already detected neighbors); the second component of the score, instead, depends on the adopted technique: for template matching methods we use the correlation score while for approaches based on local features  we rely on the number of correct matches associated with the proposal. Both terms are normalized to 1 and averaged out to get the final score assigned to  each Detection Proposal.
Based on such a score, we pick the best proposal and add it to the \textit{Observed Planogram}, so as to enforce new constraints that may be deployed throughout successive iterations to select the best-constrained missing item as well as improve ROI localization. If either all detection proposals are discarded due to the overlap check or the best one exhibits too low a score, our pipeline reports a planogram compliance issue related to the currently analyzed missing product. We have not investigated yet on how to disambiguate between different issues such as low in stock items and misplaced items. In real settings, however, such different issues would both be dealt with by manual intervention of sales clerks.  The iterative procedure stops when all the seemingly missing products have been either detected or labeled as compliance issues. 

\section{Experimental Results}
\label{sec:test}

\begin{figure}[t]
\centering
\includegraphics[width=80mm]{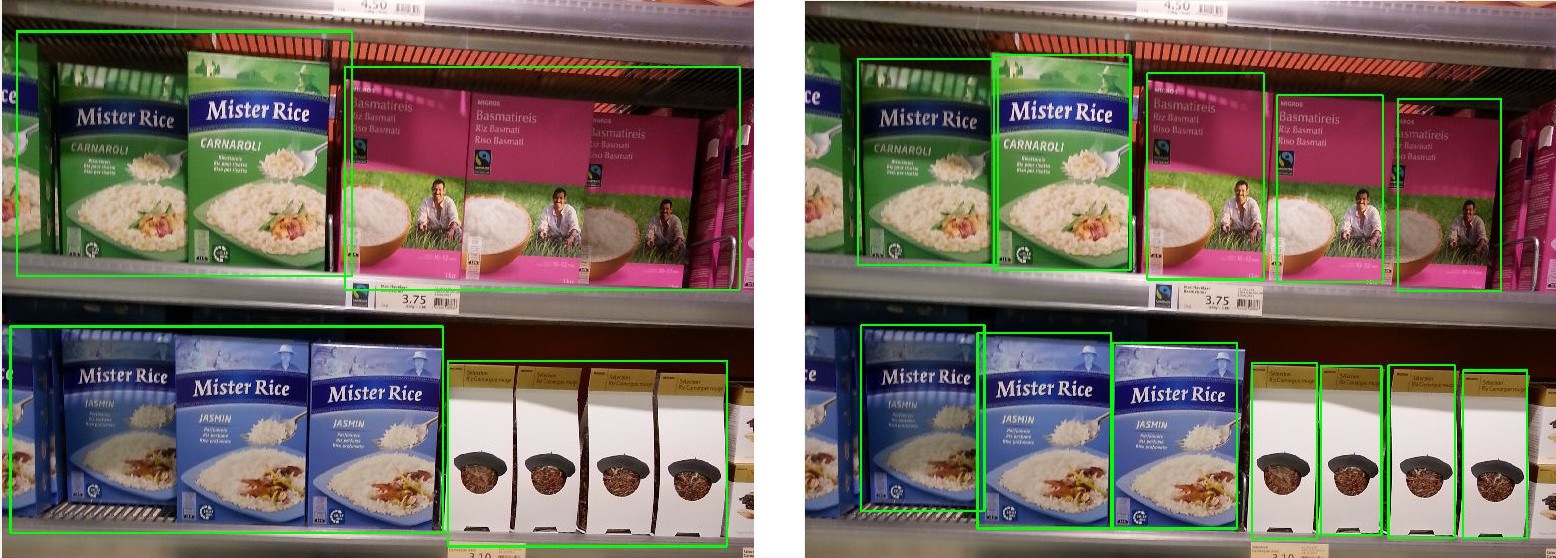} 
\caption{Ground-truth dealing with product types provided with the \textit{Grocery Products} dataset(left) and our instance-specific bounding boxes (right).}
\label{fig:bb}
\end{figure}
 
To assess the performance of our pipeline we rely on the  \textit{Grocery Products} dataset \cite{george2014recognizing}. However, as the ground-truth available with shelves images concerns product types while we aim at detecting each individual instance, we have manually annotated a subset of images with item-specific bounding boxes (see \autoref{fig:bb}). 
Moreover, for each image we have created an ideal planogram encoded in our graph like representation for the perfect disposition of products (e.g. if the actual image contains voids or misplaced items they will not be encoded on the ideal planogram that instead will model only the correct product disposition). The annotation used are available at our project page \footnote{\url{vision.disi.unibo.it/index.php?option=com_content&view=article&id=111&catid=78}}.

Our chosen subset consists of 70 images featuring box-like packages and dealing with different products such as rice, coffee, cereals, tea, juices, biscuits\dots. Each image depicts many visible products, for a total of 872 instances of 181 different products, that is on average $\approx$ 12 instances per image. According to the metric used in the PASCAL VOC challenge, we judge a detection as correct if the intersection over union between the detected and ground-truth bounding boxes  is $>0.5$. For each image we compute \emph{Precision} (number of correct detections over total number of detections), \emph{Recall} (number of correctly detected products over number of products visible in the image) and  \emph{F-Measure} (harmonic mean of \emph{Precision} and \emph{Recall}). Then, we provide charts reporting average figures across the dataset.

\begin{figure}[t]
\centering
\includegraphics[width=80mm]{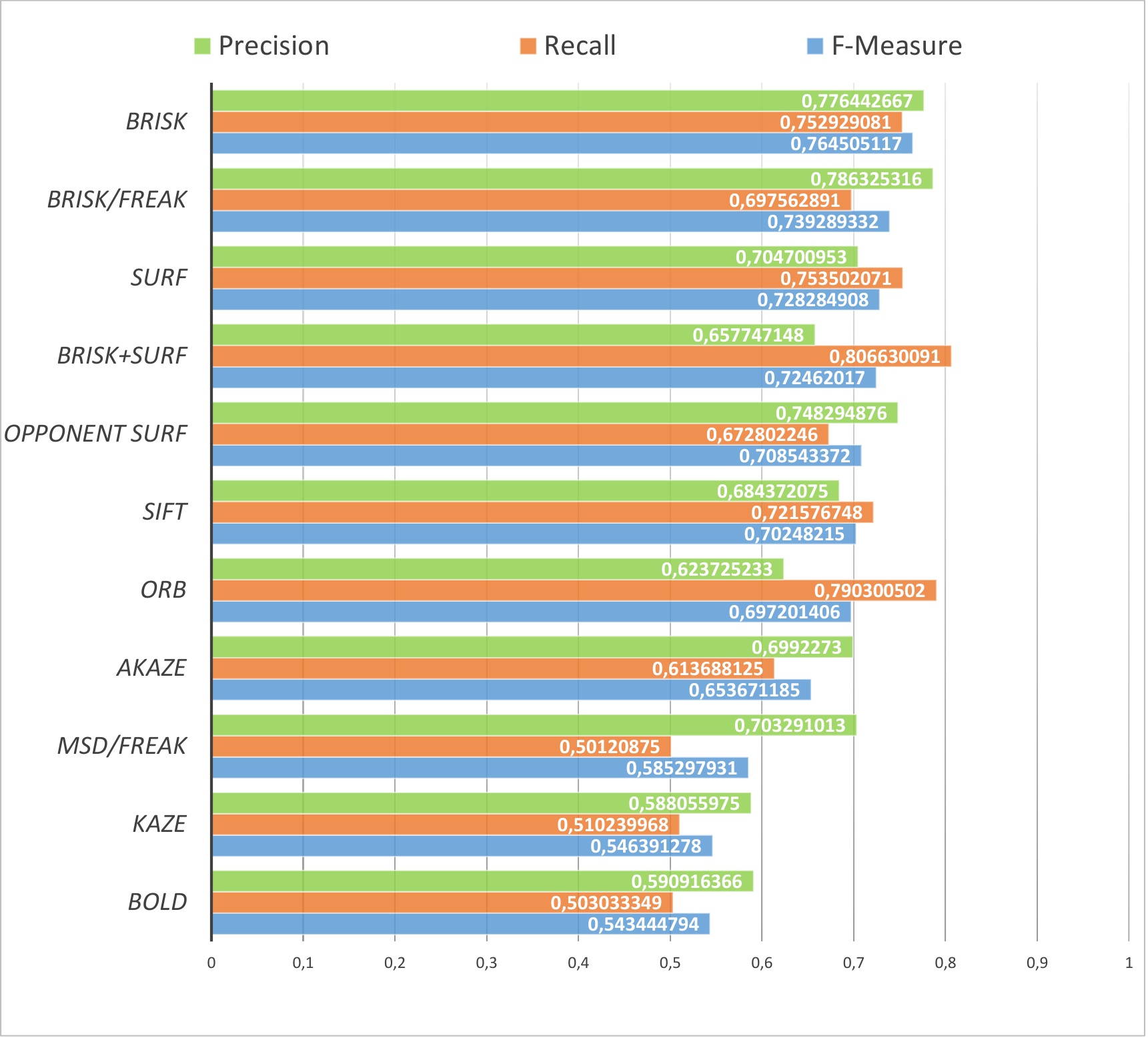} 
\caption{Evaluation of different features for \textbf{Unconstrained Product Recognition}. Results ordered from top to bottom alongside with \emph{F-Measure} scores.}
\label{fig:test1}
\end{figure}

\begin{figure}[t]
\centering
\includegraphics[width=80mm]{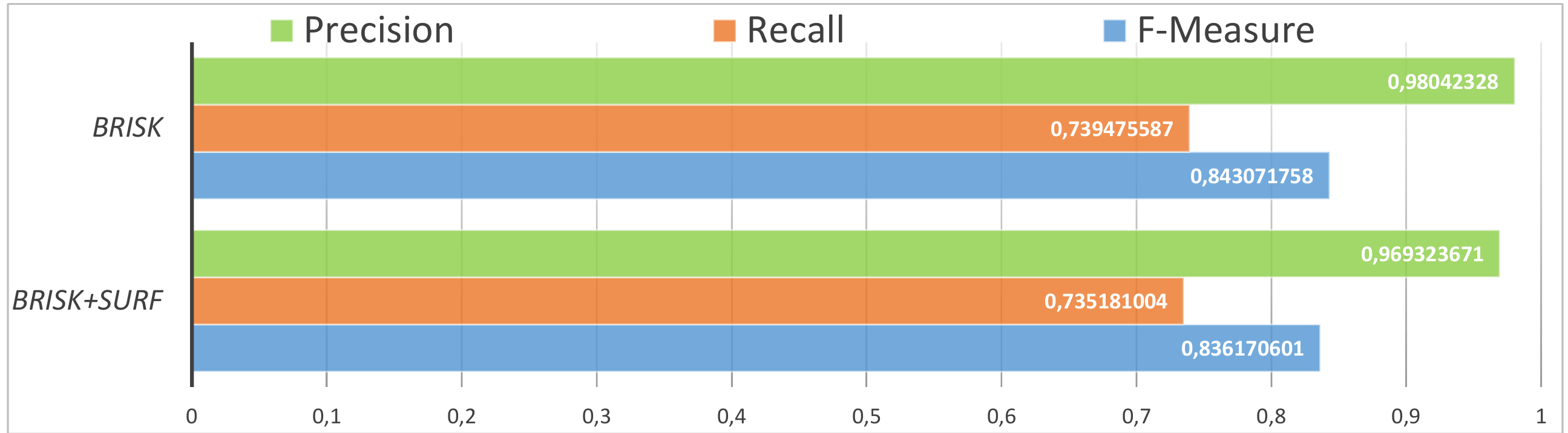} 
\caption{Results after \textbf{Graph-based Consistency Check} when using either BRISK or BRISK+SURF in the first step. }
\label{fig:test_pipe2}
\end{figure}

\begin{figure}[t]
\centering
\includegraphics[width=80mm]{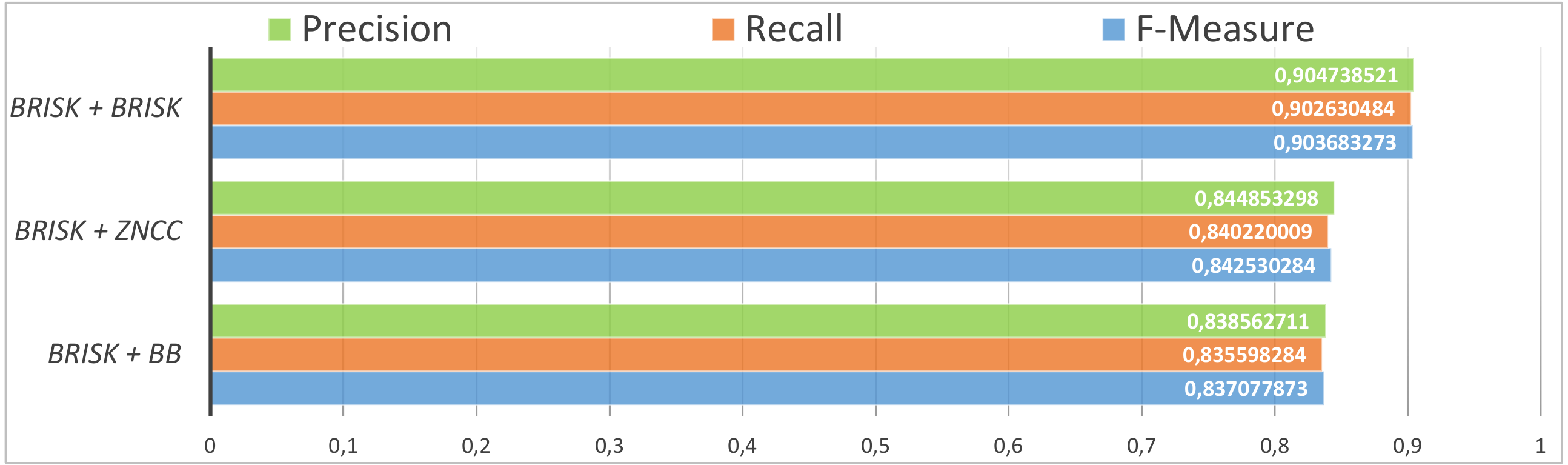} 
\caption{Evaluation of different choices for the final \textbf{Product Verification} step of our pipeline, with BRISK features used in the first step.}
\label{fig:test_pipe3}
\end{figure}

\begin{figure*}[t]
\centering
\includegraphics[width=55mm]{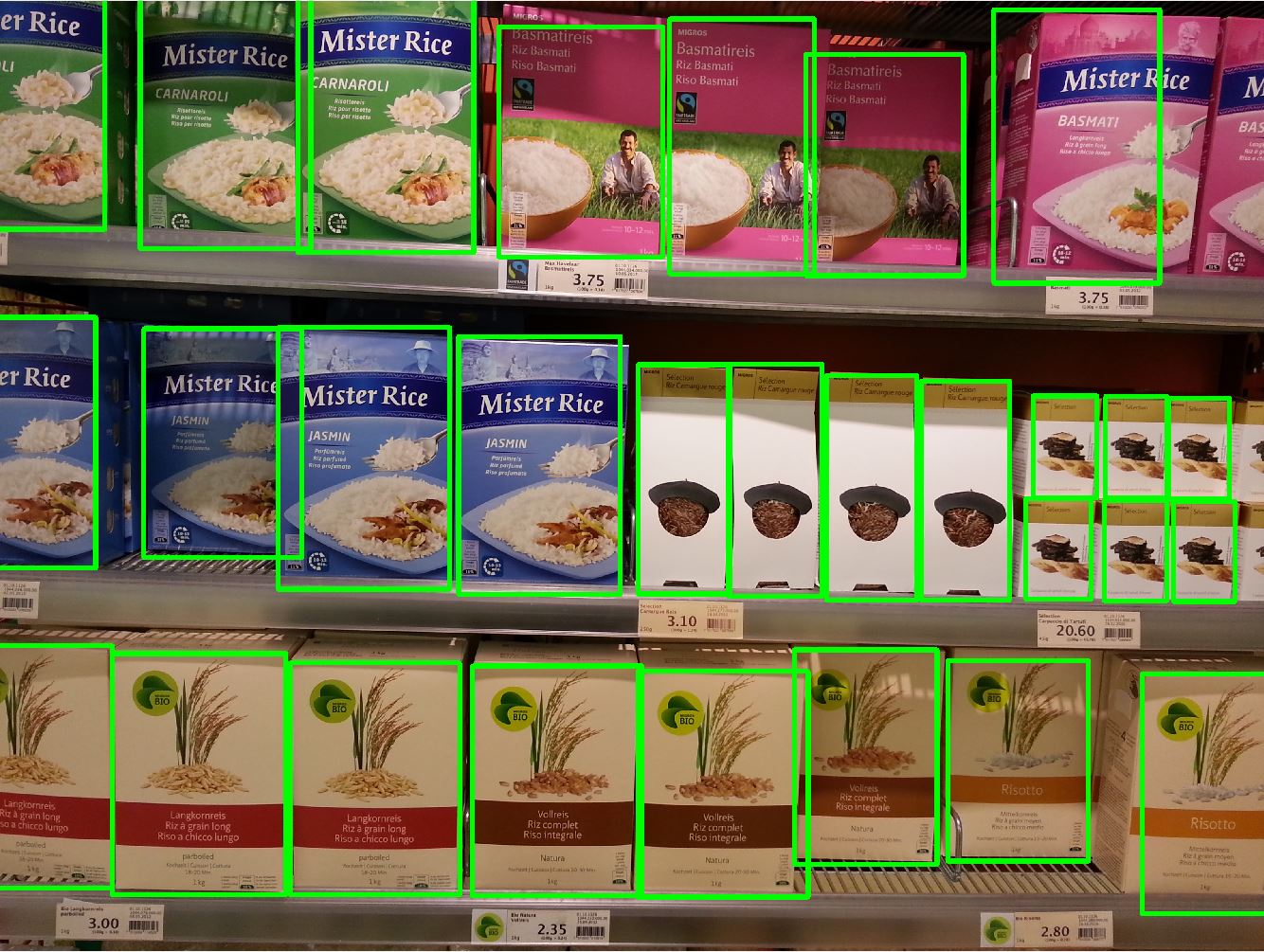} 
\includegraphics[width=55mm]{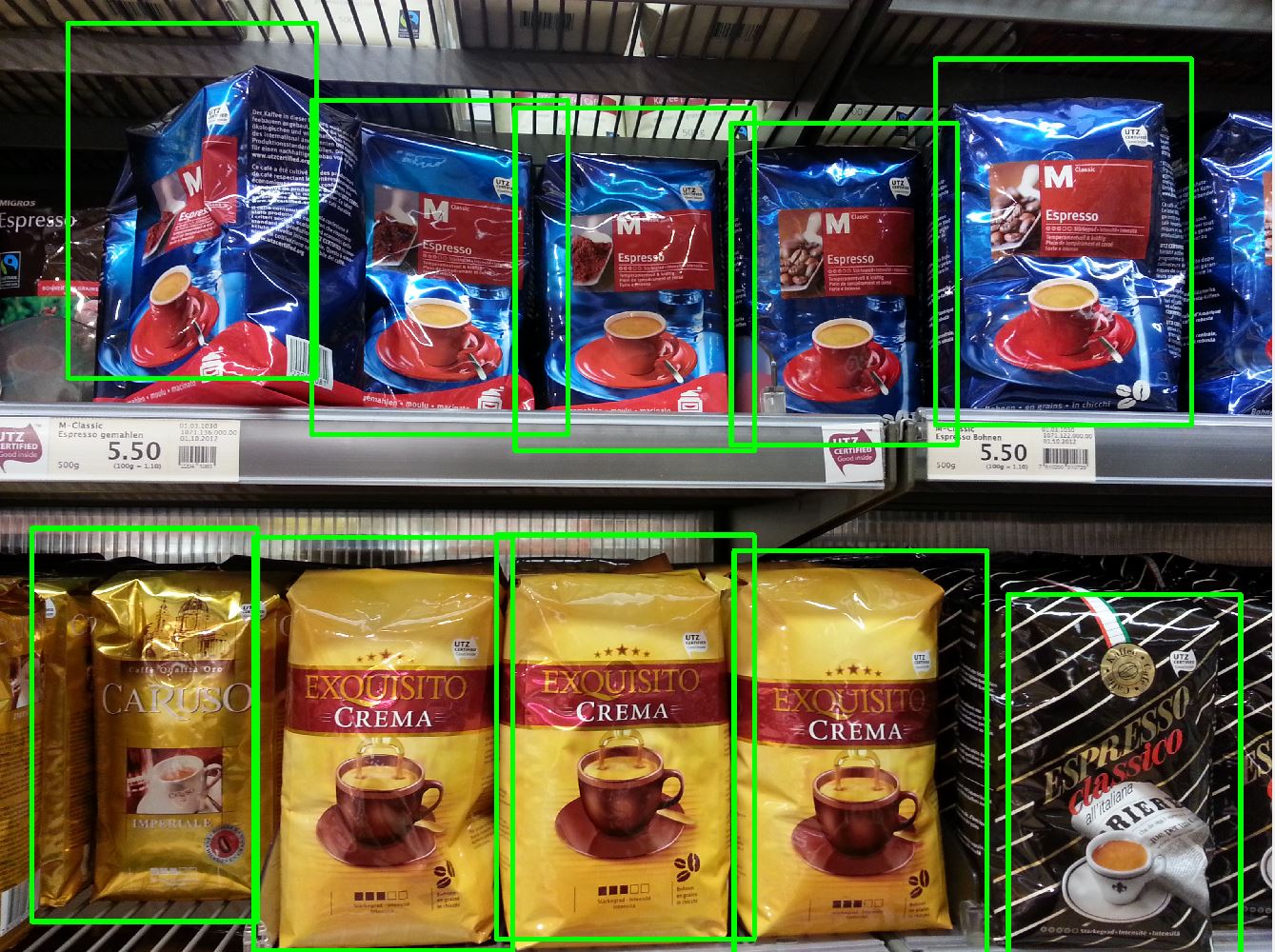} 
\includegraphics[width=55mm]{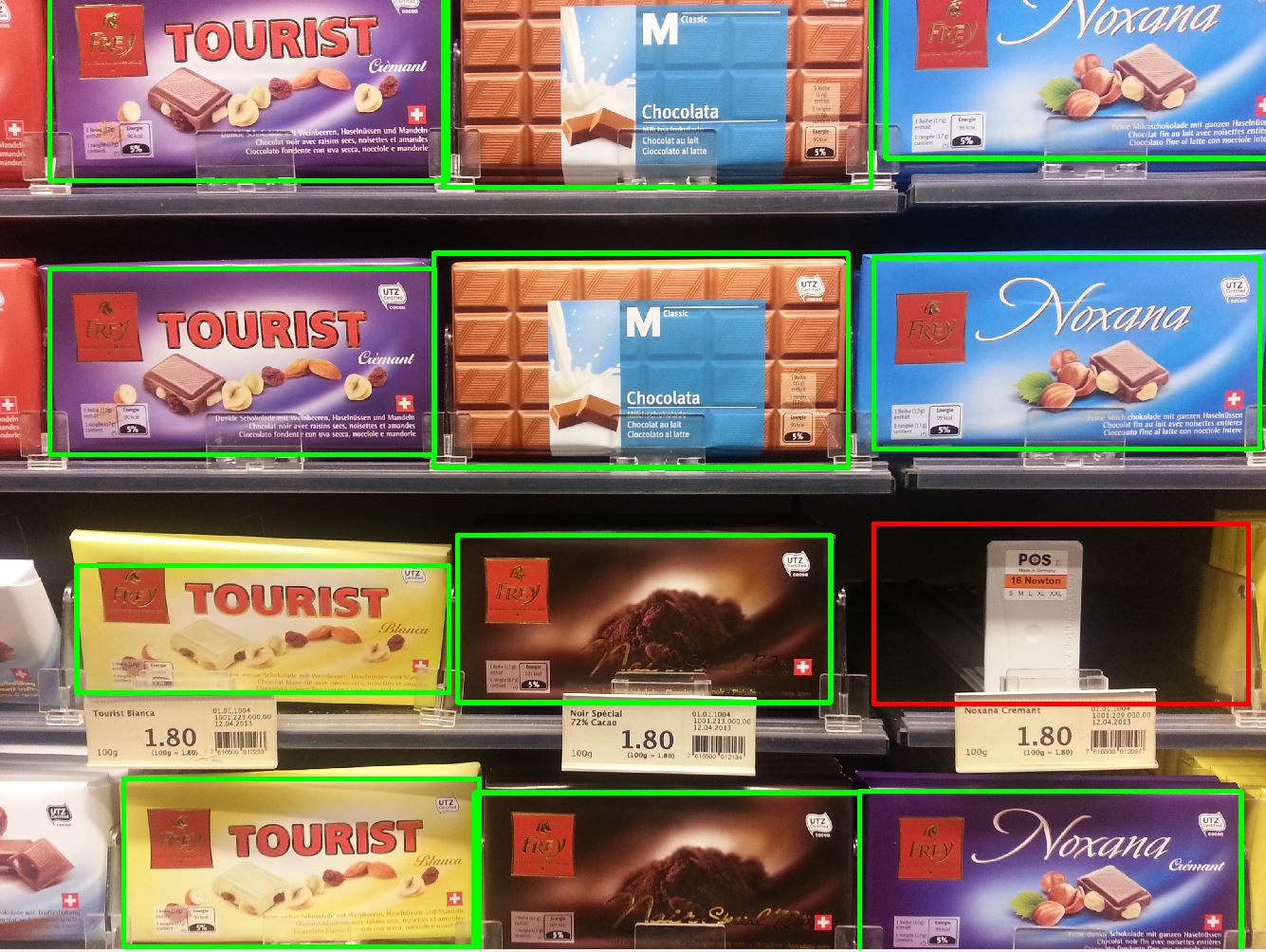} 
\caption{Qualitative results obtained by our pipeline: detected items are enclosed by green boxes while planogram compliance issues are highlighted by red boxes.}
\label{fig:qualitativo}
\end{figure*}

As regards comparative evaluation with respect to previous work, it is worth highlighting that the only work addressing exactly the same task as ours is \cite{marder2015using}, but neither their dataset nor their implementation are publicly available. Indeed, their system is quite complex and tailored for their specific use case so it would have been unfair to reproduce their results on our dataset by our own implementation based only on their paper. 
We have also investigated on the use of region proposals, such as \cite{uijlings2013selective}, followed by classification (e.g. by a CNN \cite{girshick2014rich}) of the identified regions image segments to pursue product recognition. Unfortunately, we found that this approach does not suit to the addressed task because in such a highly textured environment proposals tend to isolate logos and very colorful details from the underlying boxes while joining similarly colored regions belonging to different nearby products. By no means, thus, the region proposals provided by state-of-the-art methods employed for object detection can provide  correct segmentations of the individual products placed on store shelves. Therefore, we think that the most reasonable baseline to compare with is given by the first step of our pipeline, i.e. the standard object instance recognition approach based on local invariant features that has been proven to work effectively in a variety of diverse premises. 

In the following we will follow  the processing flow along our pipeline so as to evaluate performance gain upon execution of each step, showing how casting the planogram compliance problem as a subgraph isomorphism can dramatically improve the performance with respect to a standard feature based pipeline.
We start with evaluating  the \textbf{Unconstrained Product Recognition} step, in order to find the best suitable local features to be used in this scenario. 

We have tested all the detectors and descriptors available in OpenCV
, i.e. SIFT\cite{lowe2004distinctive}, SURF\cite{bay2006surf}, ORB\cite{rublee2011orb}, BRISK\cite{leutenegger2011brisk}, KAZE\cite{alcantarilla2012kaze}, AKAZE\cite{alcantarilla2011fast}, STAR\cite{agrawal2008censure},  MSD\cite{tombari2014interest}, FREAK\cite{alahi2012freak}, DAISY\cite{tola2010daisy}, LATCH\cite{levi2015latch}, Opponent Color Space Descriptors\cite{van2008color}, as well as the the line segments features known as BOLD\cite{tombari2013bold}(original code distributed by the authors for research purposes). We have considered features providing both the detector and descriptor (e.g. SIFT) as well as many different detector/descriptor pairs (e.g. MSD/FREAK) and multiple feature processes voting altogether in the same pose space (e.g. BRISK+SURF). A summary of the best results is reported in Figure\autoref{fig:test1}. As it can be observed, binary descriptors, such as BRISK and FREAK performs fairly well in the addressed product recognition scenario, yielding the highest \emph{Precision} and best \emph{F-Measure} scores. SURF features provide good results alike, in particular as concerns \emph{Recall}. It is also worth noticing how the use of multiple features, such as BRISK + SURF,  to capture different image structures may help increasing the sensitivity of the pipeline, as vouched by the highest \emph{Recall}. ORB features may yield a comparably high \emph{Recall}, but at expense of a lower \emph{Precision}.  The use of color descriptors (Opponent SURF), instead, does not seem to provide significant benefits. As the second step is meant to prune out the false detections provided by the first, one would be lead to prefer those features yielding higher \emph{Recall}. Yet, it may turn out hard for the second step to  solve the \emph{sub-graph isomorphism} problem in presence of too many false positives. Thus, a good balance between the two types of detection errors turns out preferable, rather.  As such, we will consider both BRISK and BRISK+SURF features within the \textbf{Unconstrained Product Recognition} step in order to further evaluate the results provided by our pipeline after the \textbf{Graph-based Consistency Check} step. 

For the second step we fixed $\tau=0.25$ and deployed the algorithm proposed in \autoref{sec:graph_consistency},
the results are displayed in \autoref{fig:test_pipe2}. First, the boost in \emph{Precision} attained with both types of features compared to the output provided by the first step (\autoref{fig:test1}) proves that the proposed \emph{sub-graph isomorphism} formulation described in \autoref{sec:graph_consistency} is very effective in robustifying product recognition by removing false detections arising in unconstrained settings. In particular, when using BRISK features, \emph{Precision} raises from $\approx$ 78\%  to $\approx$ 98\% and with BRISK+SURF from $\approx$ 66\% to $\approx$ 97\%.
Alongside, though, we observe a decrease in \emph{Recall}, such as from $\approx$ 75\% to $\approx$ 74\% with BRISK and from $\approx$ 81\% to $\approx$ 74\% with BRISK+SURF. This is mostly due to items that, although detected correctly in the first step, cannot rely on enough self-coherent neighbors to be validated 
(i.e $c(n_I,n_O)<\tau$).
 Overall, the \textbf{Graph-based Consistency Check} does improves performance significantly, as the \emph{F-Measure} increases from $\approx$ 76\% to $\approx$ 84\% and from $\approx$ 72\% to $\approx$ 84\% with BRISK and BRISK+SURF, respectively. 

Given that in \autoref{fig:test_pipe2}  BRISK slightly outperforms BRISK+SURF according to all the performance indexes and requires less computation, we pick the former features for the fist step and evaluate different design choices as regards the final \textbf{Product Verification}. In particular, as mentioned in \autoref{sec:localSearch}, we considered different template matching and feature-based approaches. The best results, summarized in \autoref{fig:test_pipe3}, concern template matching by the ZNCC (Zero-mean Normalized Cross Correlation) in the HSV color space, the recent \emph{Best-buddies Similarity} method \cite{dekel2015best} in the RGB color space and a feature-based approach deploying the same features as in the first step, that is BRISK.  As shown in \autoref{fig:test_pipe3}, using BRISK features in both the first and last step does provide the best results, all the three performance indexes getting now as high as $\approx$ 90\%.

Eventually, as for computational efficieny, our system takes at most ~15 sec per shelve image with single thread execution on a laptop PC, of which ~1 sec is spent searching for the subgraph isomorphism.
Eventually, in \autoref{fig:qualitativo} we present some qualitative results obtained  by our pipeline both in case of compliance between the observed scene and the planogram as well as in the case of missing products. Additional qualitative results are provided with the supplementary material.


\section{Conclusion and Future Work}
\label{sec:conclusioni}
We have shown how deploying product arrangement constraints by an original formulation of the product recognition problem as a sub-graph isomorphism can improve performance  dramatically compared to an unconstrained formulation. Accordingly, our proposed pipeline can work effectively in realistic scenarios in which just one model image per product and the planogram are  available and the given image is not a priory localized with respect to the aisle. 
Unfortunately, a  quantitative comparison to the most relevant previous work \cite{marder2015using} is not feasible, as the authors used a dataset that cannot be make public. Nonetheless, their dataset seems comparable to ours in terms of number of different products and instances. We report a higher recognition rate (\emph{Recall}), i.e. 90.2\% vs 87.4 \%,  with a (\emph{Precision}) as good as 90.4 \%. To enable reproducibility of results and foster future work on the topic of product recognition for planogram compliance we will made our annotated dataset public through our project's website. 

Our pipeline works quite well when applied to textured piece-wise planar products. However, grocery stores and supermarkets usually sells many different categories of products, such as bottles, jars, deformable items or even texture-less objects, like e.g. kitchenware, for which local invariant features are likely to fail in providing enough unconstrained detections to build a reliable \emph{Observed Planogram}. To address this more challenging scenario, we plan to devise a preliminary product categorization step based on machine (deep) learning to segment the image into regions corresponding to different categories (e.g. piece-wise planar packages, bottles, jars, cans kitchenware..). Purposely, we plan to rely on a similar graph-based formulation to deploy known arrangement constraints (e.g. cans are below jars). Then, each detected segment may be handled by a specific approach to establish upon planogram compliance, the method described in this paper being  applicable within segments labeled as piece-wise planar products.

\newpage
{\small
\bibliographystyle{ieee}
\bibliography{biblio}

\begin{thebibliography}{10}\itemsep=-1pt

\bibitem{advani2015visual}
S.~Advani, B.~Smith, Y.~Tanabe, K.~Irick, M.~Cotter, J.~Sampson, and
  V.~Narayanan.
\newblock Visual co-occurrence network: using context for large-scale object
  recognition in retail.
\newblock In {\em Embedded Systems For Real-time Multimedia (ESTIMedia), 2015
  13th IEEE Symposium on}, pages 1--10. IEEE, 2015.

\bibitem{agrawal2008censure}
M.~Agrawal, K.~Konolige, and M.~R. Blas.
\newblock Censure: Center surround extremas for realtime feature detection and
  matching.
\newblock In {\em Computer Vision--ECCV 2008}, pages 102--115. Springer, 2008.

\bibitem{alahi2012freak}
A.~Alahi, R.~Ortiz, and P.~Vandergheynst.
\newblock Freak: Fast retina keypoint.
\newblock In {\em Computer Vision and Pattern Recognition (CVPR), 2012 IEEE
  Conference on}, pages 510--517. Ieee, 2012.

\bibitem{alcantarilla2012kaze}
P.~F. Alcantarilla, A.~Bartoli, and A.~J. Davison.
\newblock Kaze features.
\newblock In {\em Computer Vision--ECCV 2012}, pages 214--227. Springer, 2012.

\bibitem{alcantarilla2011fast}
P.~F. Alcantarilla and T.~Solutions.
\newblock Fast explicit diffusion for accelerated features in nonlinear scale
  spaces.
\newblock {\em IEEE Trans. Patt. Anal. Mach. Intell}, 34(7):1281--1298, 2011.

\bibitem{bay2006surf}
H.~Bay, T.~Tuytelaars, and L.~Van~Gool.
\newblock Surf: Speeded up robust features.
\newblock In {\em Computer vision--ECCV 2006}, pages 404--417. Springer, 2006.

\bibitem{cotter2014hardware}
M.~Cotter, S.~Advani, J.~Sampson, K.~Irick, and V.~Narayanan.
\newblock A hardware accelerated multilevel visual classifier for embedded
  visual-assist systems.
\newblock In {\em Proceedings of the 2014 IEEE/ACM International Conference on
  Computer-Aided Design}, pages 96--100. IEEE Press, 2014.

\bibitem{dekel2015best}
T.~Dekel, S.~Oron, M.~Rubinstein, S.~Avidan, and W.~T. Freeman.
\newblock Best-buddies similarity for robust template matching.
\newblock In {\em Computer Vision and Pattern Recognition (CVPR), 2015 IEEE
  Conference on}, pages 2021--2029. IEEE, 2015.

\bibitem{frontoni2015embedded}
E.~Frontoni, A.~Mancini, and P.~Zingaretti.
\newblock Embedded vision sensor network for planogram maintenance in retail
  environments.
\newblock {\em Sensors}, 15(9):21114--21133, 2015.

\bibitem{frontoni2014information}
E.~Frontoni, A.~Mancini, P.~Zingaretti, and V.~Placidi.
\newblock Information management for intelligent retail environment: The shelf
  detector system.
\newblock {\em Information}, 5(2):255--271, 2014.

\bibitem{george2014recognizing}
M.~George and C.~Floerkemeier.
\newblock Recognizing products: A per-exemplar multi-label image classification
  approach.
\newblock In {\em Computer Vision--ECCV 2014}, pages 440--455. Springer, 2014.

\bibitem{girshick2014rich}
R.~Girshick, J.~Donahue, T.~Darrell, and J.~Malik.
\newblock Rich feature hierarchies for accurate object detection and semantic
  segmentation.
\newblock In {\em Proceedings of the IEEE conference on computer vision and
  pattern recognition}, pages 580--587, 2014.

\bibitem{gruen2002retail}
T.~W. Gruen, D.~Corsten, and S.~Bharadwaj.
\newblock Retail out of stocks: A worldwide examination of causes, rates, and
  consumer responses.
\newblock {\em Grocery Manufacturers of America, Washington, DC}, 2002.

\bibitem{larsen2013automated}
B.~J. Larsen.
\newblock Automated generation of a three-dimensional space representation and
  planogram verification, Nov.~6 2013.
\newblock US Patent App. 14/073,231.

\bibitem{leutenegger2011brisk}
S.~Leutenegger, M.~Chli, and R.~Y. Siegwart.
\newblock Brisk: Binary robust invariant scalable keypoints.
\newblock In {\em Computer Vision (ICCV), 2011 IEEE International Conference
  on}, pages 2548--2555. IEEE, 2011.

\bibitem{levi2015latch}
G.~Levi and T.~Hassner.
\newblock Latch: Learned arrangements of three patch codes.
\newblock {\em arXiv preprint arXiv:1501.03719}, 2015.

\bibitem{lowe2004distinctive}
D.~G. Lowe.
\newblock Distinctive image features from scale-invariant keypoints.
\newblock {\em International journal of computer vision}, 60(2):91--110, 2004.

\bibitem{mankodiya2012challenges}
K.~Mankodiya, R.~Gandhi, and P.~Narasimhan.
\newblock Challenges and opportunities for embedded computing in retail
  environments.
\newblock In {\em Sensor Systems and Software}, pages 121--136. Springer, 2012.

\bibitem{marder2015using}
M.~Marder, S.~Harary, A.~Ribak, Y.~Tzur, S.~Alpert, and A.~Tzadok.
\newblock Using image analytics to monitor retail store shelves.
\newblock {\em IBM Journal of Research and Development}, 59(2/3):3--1, 2015.

\bibitem{merler2007recognizing}
M.~Merler, C.~Galleguillos, and S.~Belongie.
\newblock Recognizing groceries in situ using in vitro training data.
\newblock In {\em Computer Vision and Pattern Recognition, 2007. CVPR'07. IEEE
  Conference on}, pages 1--8. IEEE, 2007.

\bibitem{opalach2012planogram}
A.~Opalach, A.~Fano, F.~Linaker, and R.~B.~R. Groenevelt.
\newblock Planogram extraction based on image processing, May~29 2012.
\newblock US Patent 8,189,855.

\bibitem{rublee2011orb}
E.~Rublee, V.~Rabaud, K.~Konolige, and G.~Bradski.
\newblock Orb: an efficient alternative to sift or surf.
\newblock In {\em Computer Vision (ICCV), 2011 IEEE International Conference
  on}, pages 2564--2571. IEEE, 2011.

\bibitem{shapiro2009planogram}
M.~Shapiro.
\newblock Executing the best planogram.
\newblock {\em Professional Candy Buyer, Norwalk, CT, USA}, 2009.

\bibitem{tola2010daisy}
E.~Tola, V.~Lepetit, and P.~Fua.
\newblock Daisy: An efficient dense descriptor applied to wide-baseline stereo.
\newblock {\em Pattern Analysis and Machine Intelligence, IEEE Transactions
  on}, 32(5):815--830, 2010.

\bibitem{tombari2014interest}
F.~Tombari and L.~Di~Stefano.
\newblock Interest points via maximal self-dissimilarities.
\newblock In {\em Computer Vision--ACCV 2014}, pages 586--600. Springer, 2014.

\bibitem{tombari2013bold}
F.~Tombari, A.~Franchi, and L.~Stefano.
\newblock Bold features to detect texture-less objects.
\newblock In {\em Proceedings of the IEEE International Conference on Computer
  Vision}, pages 1265--1272, 2013.

\bibitem{uijlings2013selective}
J.~R. Uijlings, K.~E. van~de Sande, T.~Gevers, and A.~W. Smeulders.
\newblock Selective search for object recognition.
\newblock {\em International journal of computer vision}, 104(2):154--171,
  2013.

\bibitem{ullmann2010bit}
J.~R. Ullmann.
\newblock Bit-vector algorithms for binary constraint satisfaction and subgraph
  isomorphism.
\newblock {\em Journal of Experimental Algorithmics (JEA)}, 15:1--6, 2010.

\bibitem{van2008color}
K.~E. van~de Sande, T.~Gevers, and C.~G. Snoek.
\newblock Color descriptors for object category recognition.
\newblock In {\em Conference on Colour in Graphics, Imaging, and Vision},
  volume 2008, pages 378--381. Society for Imaging Science and Technology,
  2008.

\bibitem{varol2015toward}
G.~Varol and R.~S. Kuzu.
\newblock Toward retail product recognition on grocery shelves.
\newblock In {\em Sixth International Conference on Graphic and Image
  Processing (ICGIP 2014)}, pages 944309--944309. International Society for
  Optics and Photonics, 2015.

\bibitem{wegener2005complexity}
I.~Wegener.
\newblock {\em Complexity theory: exploring the limits of efficient
  algorithms}.
\newblock Springer Science \& Business Media, 2005.

\bibitem{winlock2010toward}
T.~Winlock, E.~Christiansen, and S.~Belongie.
\newblock Toward real-time grocery detection for the visually impaired.
\newblock In {\em Computer Vision and Pattern Recognition Workshops (CVPRW),
  2010 IEEE Computer Society Conference on}, pages 49--56. IEEE, 2010.

\end{thebibliography}
}

\newpage


\section*{Supplementary material}

Results obtained by our pipeline, good detection are enclosed in green bounding boxes, errors in red. Each image depicts a different step: (A) *Unconstrainstrained Product Recognition,* (B) *Graph-based Consistency Check*, (C) *Product Verification.*

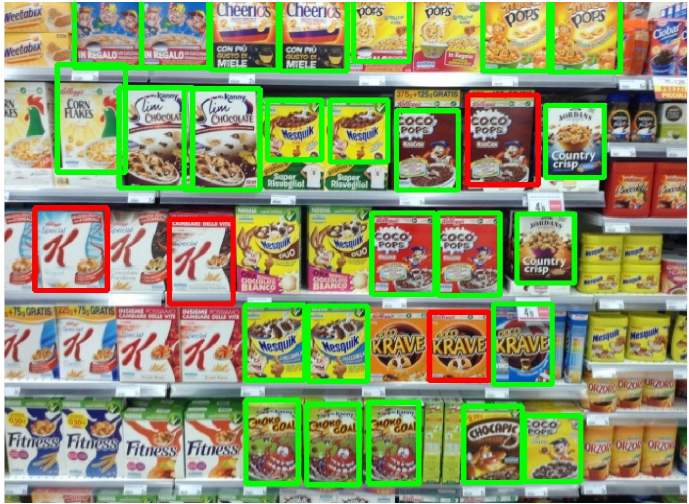
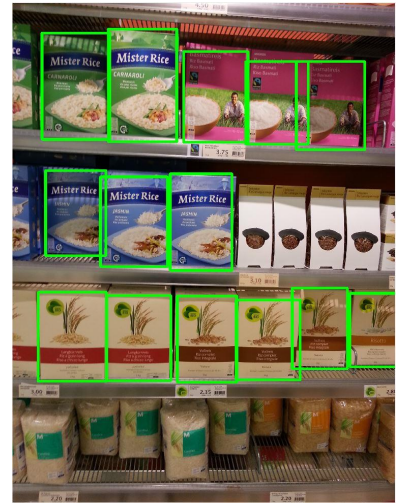

(A)

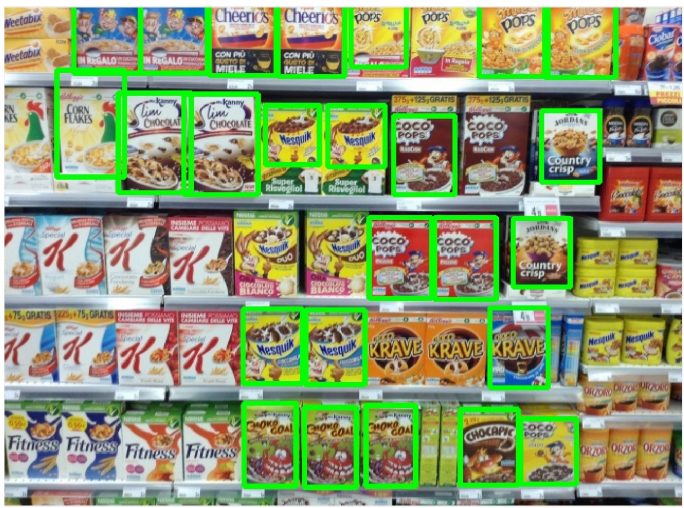
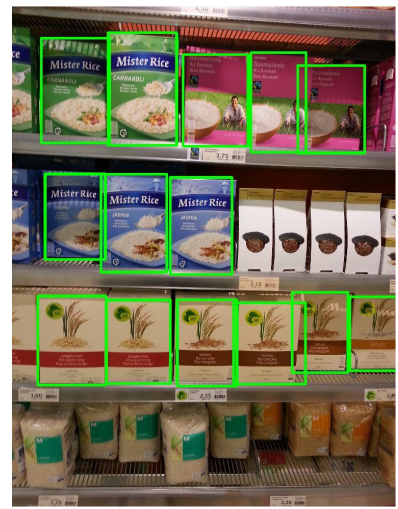

(B)

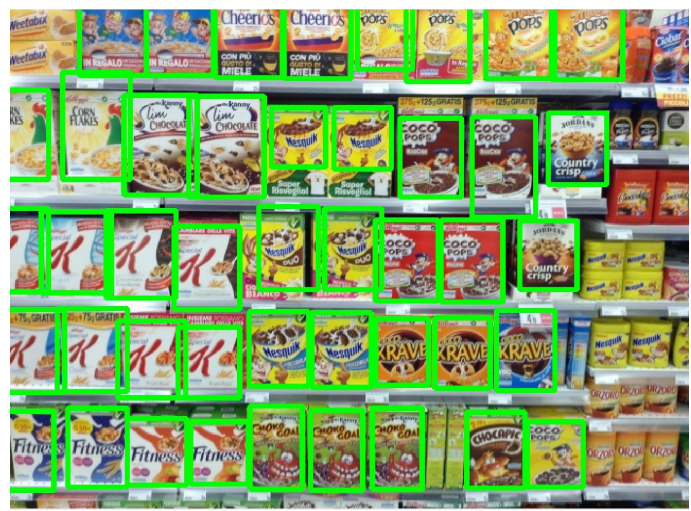
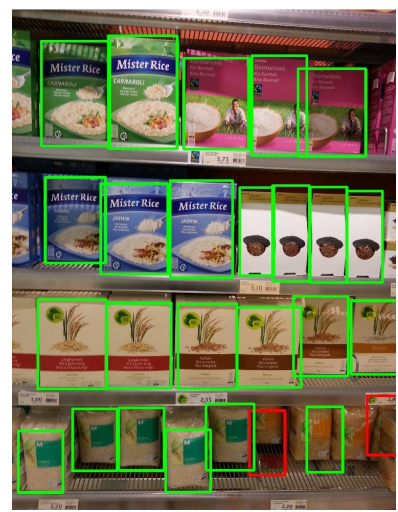

(C)

In the left column the final result shows successful detection of all products (42). Those not found do not belong to the model set. In the right column the final result contains two errors, both are indeed related to imprecise localization of the bounding boxes though the identified products are correct.

Effectiveness of the *Graph Based Consistency Check.* (A) portrays detection by *Unconstrained Product Recognition* (Green=good, Red=mistake), (B) highlights how *Graph Based Consistency Check* prunes out false detections.

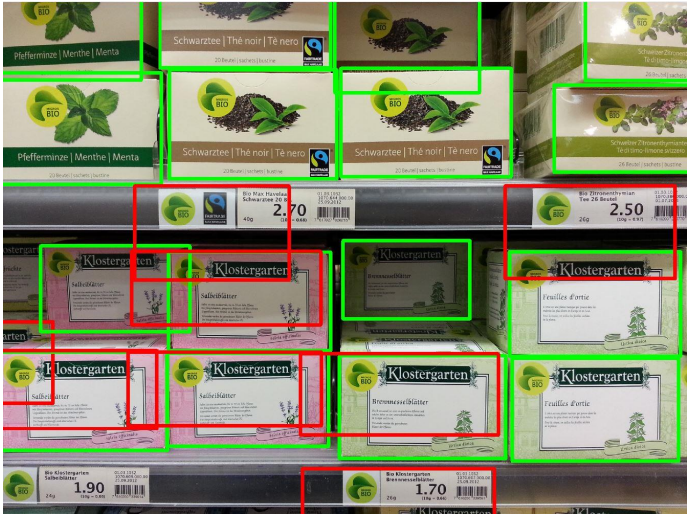
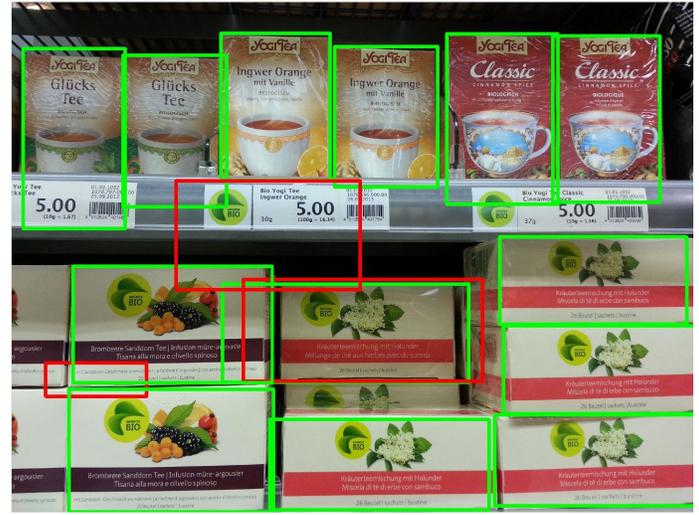

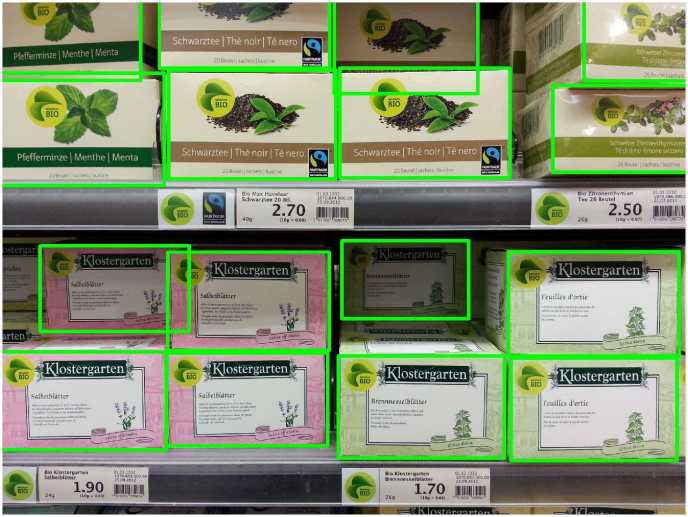
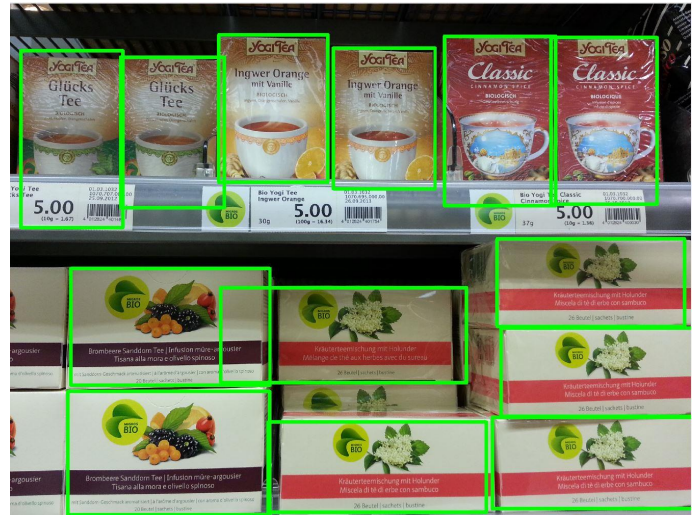

Some false detections in (A) are due to the presence of the green '*bio*' symbols both on products and next to the price tags. Our graph based check can succesfully prune out all of the errors as depicted in (B).

Detections of **planogram compliance issues** (orange boxes with a 'x' inside). As before we show the results at the three steps of our pipeline, after the first one (A) almost all the products in the scene have been detected altough there are some mistakes, the graph based check (B) remove false detections and finally the product verification step (C) completes the observed planogram and identifies potential compliance issue.

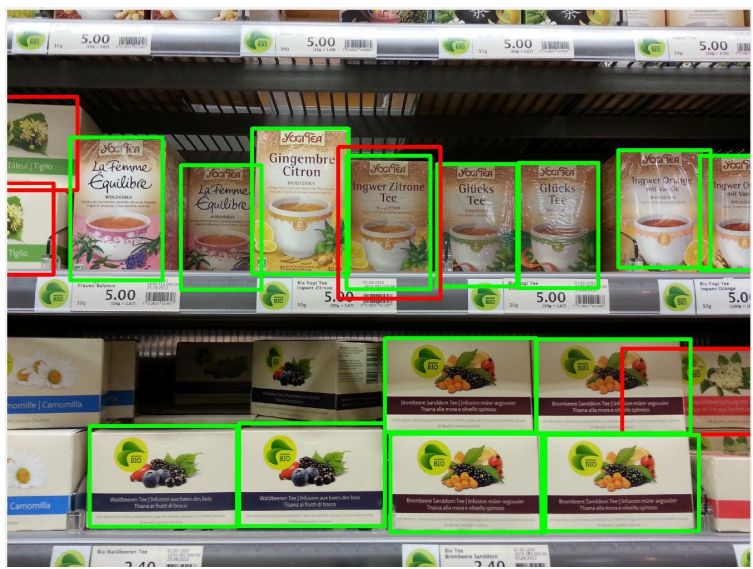
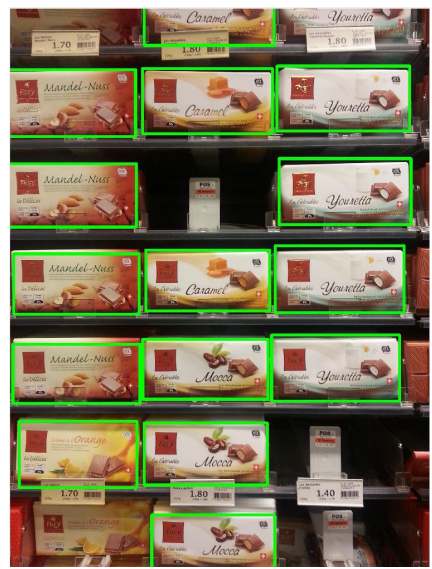
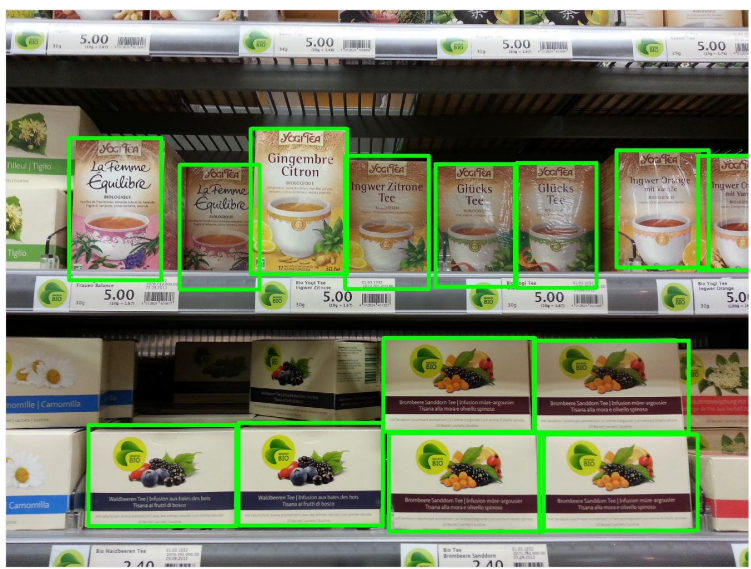
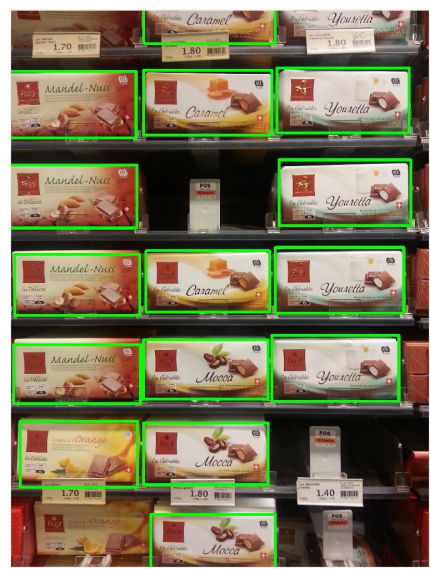
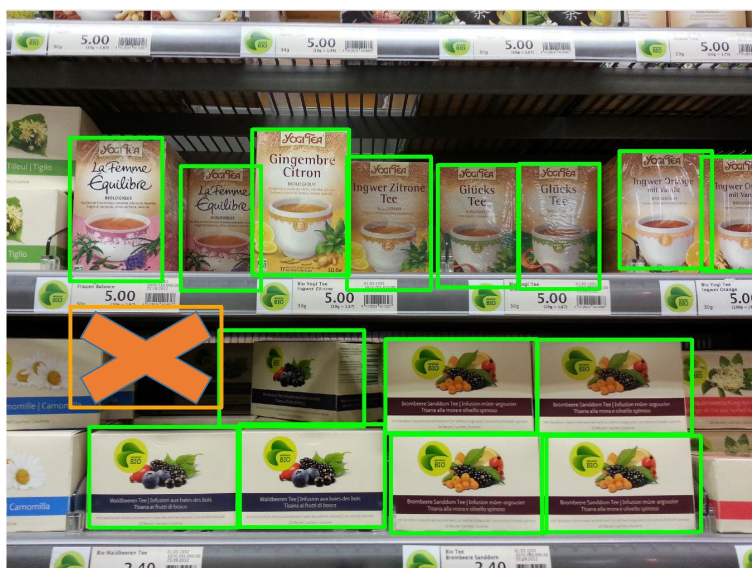
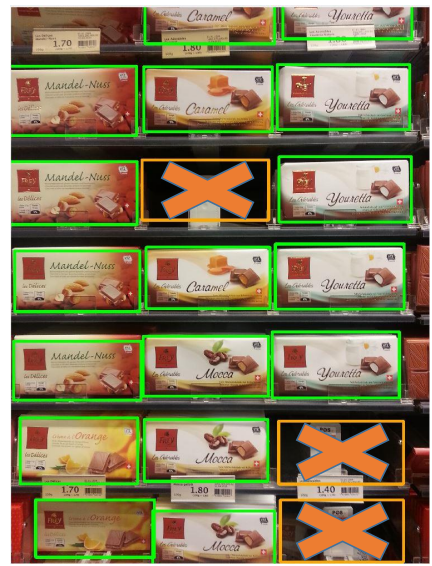

After *Unconstrained Product Detection* only 6 out of 11 products are correctly identified. 2 of the missing ones are false detections due to very similar packages. 3 are not found at all.

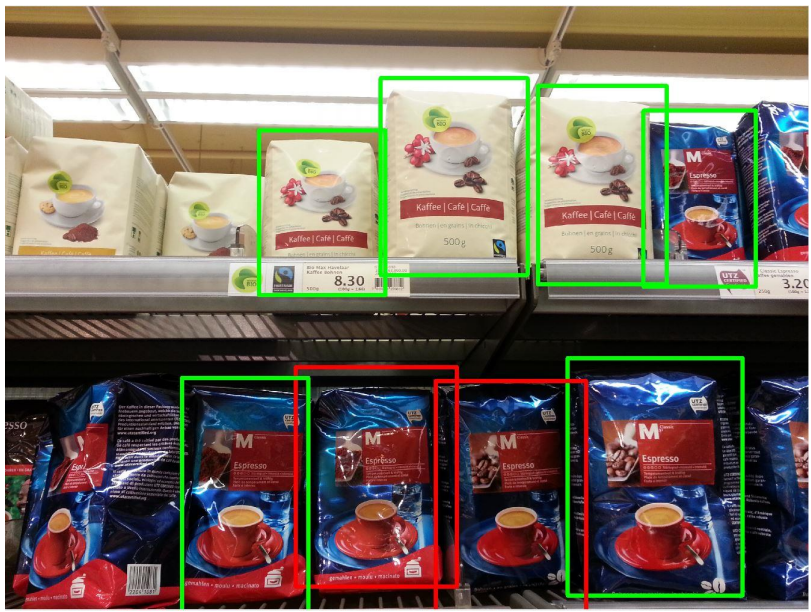

*Graph Based Consistency Check* successfully removes the two false detections, though a good detection has been pruned out too due to a low number of coherent neighbors (just one in this case).

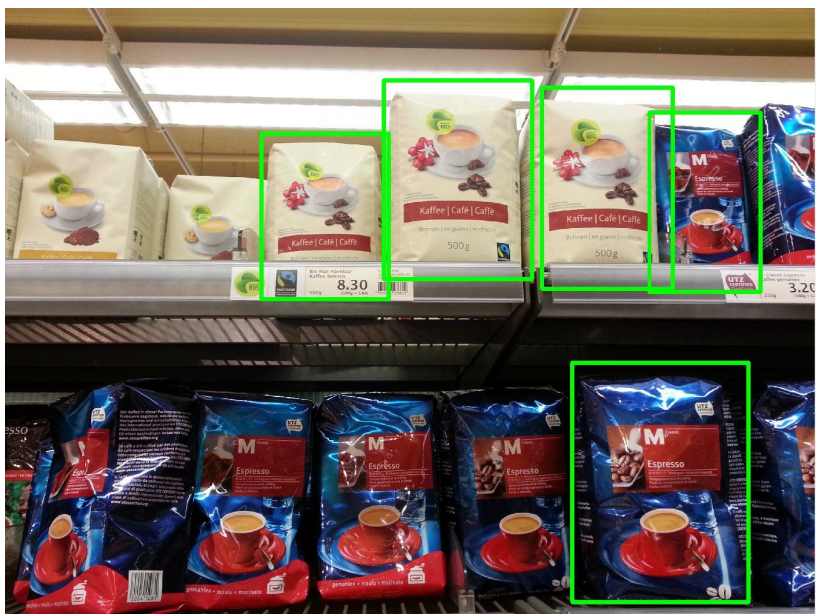

Finally, *Product Verification* completes the observed planogram by finding the 6 missing products one at a time. The partially skewed coffe package on the left has been correctly detected altough only partially visible.

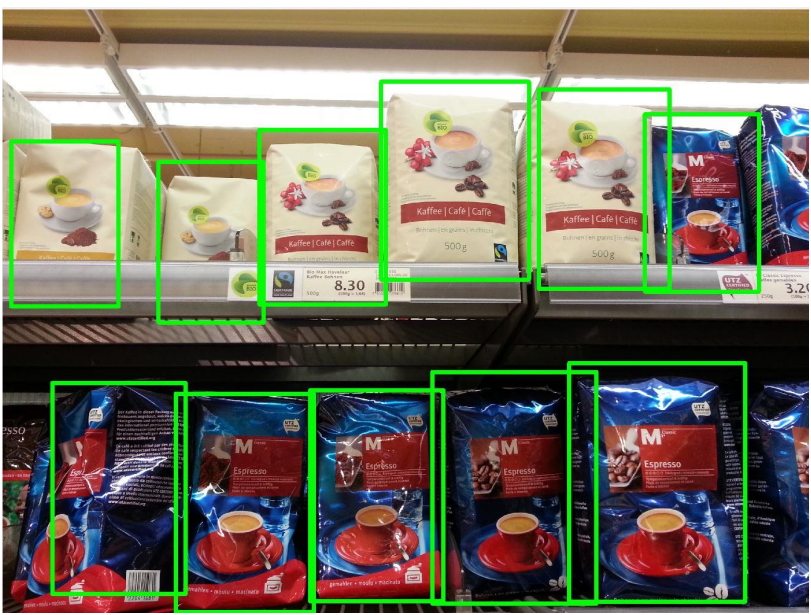



\end{document}